\documentclass[manuscript]{acmart}

\usepackage{textcomp}
\usepackage{multirow}
\usepackage{xcolor}
\usepackage[capitalise]{cleveref}
\usepackage{subcaption}
\usepackage{makecell}
\usepackage{rotating}
\usepackage{xcolor}
\usepackage{algorithm}
\usepackage{algorithmic}
\usepackage{tablefootnote}
\usepackage{ctable}
\usepackage{threeparttable}
\definecolor{Gold}{HTML}{FFB800}    % Brighter, richer gold
\definecolor{Silver}{HTML}{A8A8A8}  % Slightly darker, better contrast silver
\definecolor{Bronze}{HTML}{C06014}  % Warmer, more vivid bronze

\AtBeginDocument{%
  }

\setcopyright{acmlicensed}
\copyrightyear{2018}
\acmYear{2018}
\acmDOI{XXXXXXX.XXXXXXX}
\acmConference[Conference acronym 'XX]{Make sure to enter the correct
  conference title from your rights confirmation email}{June 03--05,
  2018}{Woodstock, NY}
\acmISBN{978-1-4503-XXXX-X/2018/06}

\begin{document}

% \title{KAN-MLP-Mixer: Improving IMU-based Human Activity Recognition with Hybrid Networks}
\title{KAN-MLP-Mixer: A comprehensive investigation of the usage of
Kolmogorov–Arnold Networks (KANs) for improving IMU-based Human Activity Recognition
}

% \title{Let KANs first: An extensive study on Leaveraging Kans for Sensor-based Human Activity Recognition}

\author{Mengxi Liu}
\email{mengxi.liu@dfki.de}
\affiliation{%
  \institution{DFKI}
  \country{Germany}
}
\author{Sizhen Bian}
\email{sizhen.bian@nwpu.edu.cn}
\affiliation{%
  \institution{Northwestern Polytechnical University}
  \country{China}
}

\author{Vitor Fortes}
\email{vitor.fortes_rey@dfki.de}
\affiliation{%
  \institution{DFKI and RPTU}
  \country{Germany}
}

\author{Francisco Calatrava Nicolas}
\affiliation{%
  \institution{Örebro University}
  \country{Sweden}
}

\author{Daniel Geißler}
\email{daniel.geissler@dfki.de}
\affiliation{%
  \institution{DFKI}
  \country{Germany}
}

\author{Maximilian Kiefer-Emmanouilidis}
\email{maximilian.kiefer@dfki.de}
\affiliation{%
  \institution{DFKI and RPTU}
  \country{Germany}
}

\author{Bo Zhou}
\email{bo.zhou@dfki.de}
\affiliation{%
  \institution{DFKI and RPTU}
  \country{Germany}
}

\author{Paul Lukowicz}
\email{paul.lukowicz@dfki.de}
\affiliation{%
  \institution{DFKI and RPTU}
  \country{Germany}
}

% \author[DFKI,RPTU]{Mengxi Liu}
% \author[DFKI,RPTU]{Sizhen Bian}
% \author[]{Francisco Calatrava Nicolas}
% \author[DFKI,RPTU]{Vitor Fortes Rey\corref{mycorrespondingauthor}}
% \author[DFKI]{Daniel Geißler}
% \cortext[mycorrespondingauthor]{Corresponding author}
% 		\ead{mengxi.liu@dfki.de}
% \author[DFKI,RPTU]{Maximilian Kiefer-Emmanouilidis}
% \author[DFKI,RPTU]{Bo Zhou}
% \author[DFKI,RPTU]{Paul Lukowicz}

% \fntext[cofirst]{These authors contributed equally to this work.}
% \address[DFKI]{German Research Center for Artificial Intelligence (DFKI), Kaiserslautern, Germany}
% \address[RPTU]{Department of Computer Science, RPTU Kaiserslautern-Landau, Kaiserslautern, Germany}

\renewcommand{\shortauthors}{Liu et al.}
\renewcommand{\shorttitle}{KAN-MLP-Mixer}

\begin{abstract}

Kolmogorov–Arnold Networks (KANs) have demonstrated an exceptional ability to learn complex functions on clean, low-dimensional data but struggle to maintain performance on noisy and imperfect real-world datasets. In contrast, conventional multi-layer perceptrons (MLPs) are far more tolerant to noise and computationally efficient. Replacing all MLP components with KANs in HAR models often degrades accuracy and computation efficiency, highlighting an open challenge: how to combine KANs’ precision with MLPs’ noise robustness and efficiency. To address this, we systematically explore various placements of KAN modules within deep HAR networks and propose a hybrid architecture that strategically synergizes the strengths of both paradigms, which uses a KAN-based input embedding layer, retains MLP layers for intermediate feature mixing, and introduces a specialized LarctanKAN module for final activity classification. Across eight public HAR datasets, the hybrid KAN–MLP model achieves an average macro F1 score relative improvement of 5.33\% compared pure-MLP model, significantly outperforming standalone KAN and MLP baselines. Furthermore, integrating this hybrid strategy into other state-of-the-art HAR architectures consistently boosts their performance. Our findings demonstrate that a carefully orchestrated combination of KAN, MLP, or other conventional neural components yields more robust and accurate HAR models for real-world wearable sensing environments.

\end{abstract}



\keywords{Do, Not, Us, This, Code, Put, the, Correct, Terms, for,
  Your, Paper}

\received{20 February 2007}
\received[revised]{12 March 2009}
\received[accepted]{5 June 2009}

\maketitle

\section{Introduction}

Accurate Human Activity Recognition (HAR) from body-worn sensors is a cornerstone of ubiquitous computing, enabling applications ranging from personalized health monitoring and fitness tracking on smartwatches \cite{abbas2024active,yin2024systematic} to context-aware interactions in smart environments \cite{abdel2021human, bian2022state}. Inertial Measurement Units (IMUs), commonly embedded in wearables, provide rich motion data. However, leveraging this data effectively is challenging; real-world IMU signals are notoriously complex and plagued by noise, sensor drift, placement variations, and inter-subject differences, hindering the development of robust, universally applicable HAR systems \cite{tseng2023hybrid}.

Deep learning models have become the standard for tackling HAR, automatically learning features from sensor data \cite{chen2021deep}. Among these, Multi-Layer Perceptrons (MLPs) and their variants remain surprisingly effective \cite{ojiako2023mlps,zhou2024mlp}. While perhaps less complex than CNNs or RNNs, MLPs are valued not only for their representational power but crucially for their \textbf{robustness to noise} and \textbf{computational efficiency} \cite{le2024exploring}. This efficiency is paramount for deployment on resource-constrained wearable devices where battery life and processing power are limited. Indeed, recent work has shown that purely MLP-based architectures, like MLP-Mixers adapted for HAR and MLPHAR, can achieve state-of-the-art or competitive performance with significantly fewer parameters than heavier models \cite{ojiako2023mlps, zhou2024mlp, miyoshi2025applying}, making them highly practical for the wearable and ubiquitous computing domain.

Recently, Kolmogorov–Arnold Networks (KANs) emerged as a novel neural network paradigm offering remarkable theoretical potential \cite{liu2024kan}. Inspired by the Kolmogorov-Arnold representation theorem, KANs replace the fixed activations and linear weights of MLPs with learnable univariate functions on network edges. This design grants them exceptional flexibility, enabling them to approximate complex functions with high fidelity, often surpassing traditional networks on clean, low-dimensional mathematical or physics-based tasks \cite{liu2024kan, drokin2024kolmogorov, poeta2024benchmarking, jamali2024learn},including new efficient architectures for quantum machine learning \cite{Werner2025, Ivashkov2026}.

\begin{figure*}[ht]
  \centering
  \begin{subfigure}[b]{0.48\textwidth}
    \includegraphics[width=\linewidth]{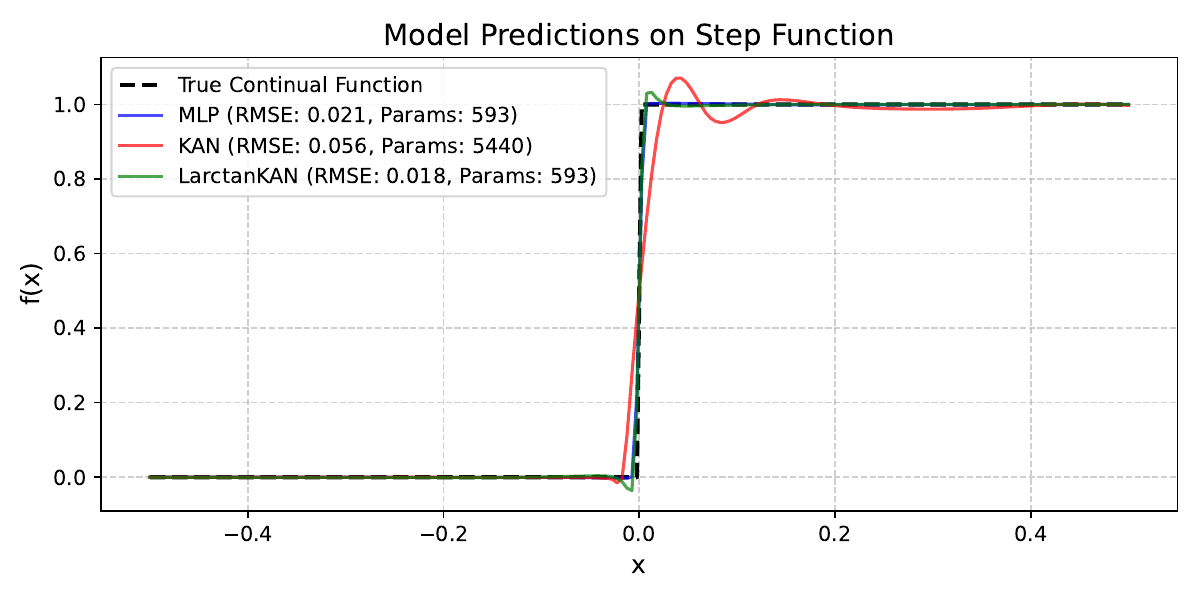}
    \caption{
      Predictions on a step function. LarctanKAN achieves the lowest error (RMSE = 0.018) with only 593 parameters, 
      outperforming both MLP and KAN, the latter of which exhibits overshooting due to its spline-based formulation 
      and uses 5440 parameters. This demonstrates the advantage of LarctanKAN for modeling discontinuous decision boundaries.
    }
    \label{fig:step-function-rmse}
  \end{subfigure}
  \hfill
  \begin{subfigure}[b]{0.48\textwidth}
    \includegraphics[width=\linewidth]{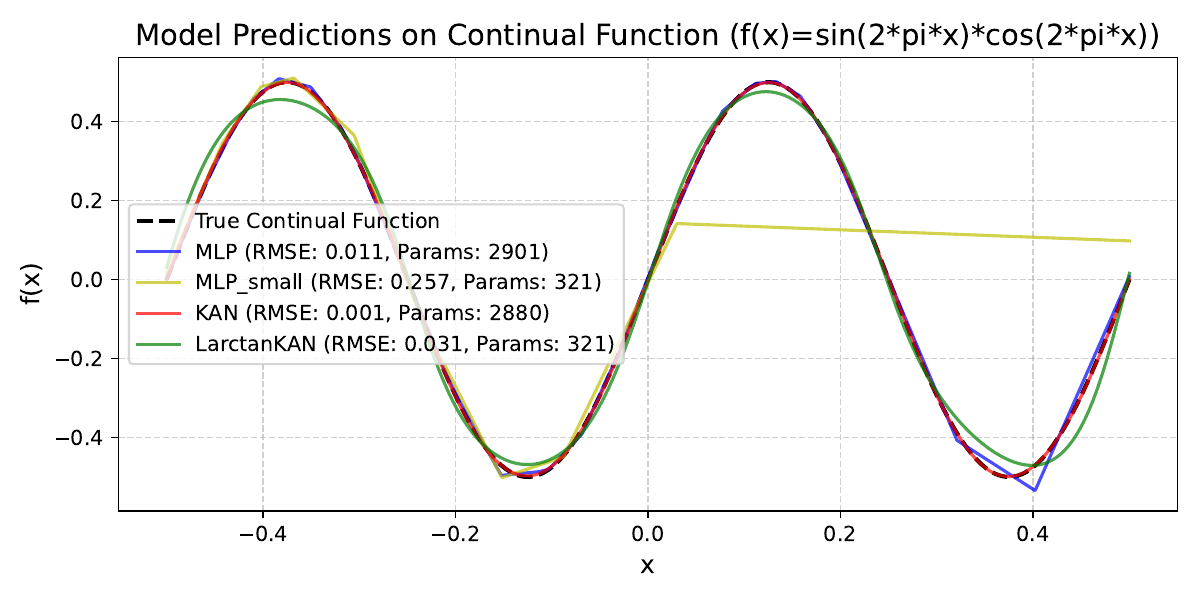}
    \caption{
      Predictions on a smooth periodic function \( f(x) = \sin(2\pi x)\cos(2\pi x) \). 
      KAN achieves the lowest RMSE (0.001), capturing the waveform precisely. 
      MLP and LarctanKAN underfit due to limited expressivity,  
      highlighting KAN's strength in embedding smooth sensor signals.
    }
    \label{fig:continual-function-rmse}
  \end{subfigure}
  \caption{
    Comparison of model predictions on synthetic functions representing typical characteristics of sensor data. 
    The step function emulates classification boundaries, while the periodic function simulates continuous IMU signal patterns.
    Results highlight the motivation for our hybrid architecture: using KAN in the embedding layer and LarctanKAN in the classifier.
  }
  \label{fig:rmse-function-comparison}
\end{figure*}

However, translating KANs' theoretical promise to the messy reality of wearable sensor data presents a significant hurdle. A growing body of evidence indicates that KANs exhibit considerable \textbf{sensitivity to noise and data imperfections} \cite{shen2025reduced, cang2024can, ibrahum2024resilient}. Their intricate function-fitting mechanism, while powerful on clean signals, appears vulnerable to the inherent variability and noise found in IMU data collected in the wild. 
Moreover, KANs are based on the assumption that the target function is continuous; if this assumption is violated, the model may fail to approximate the function effectively \cite{liu2024kan}, as shown in \cref{fig:rmse-function-comparison}. As we can see there, KANs can fit better smooth periodic signals while MLPs outperform them in decision boundaries such as step functions.
Directly substituting KANs for MLPs in established HAR pipelines often results in a substantial drop in accuracy and can negate the efficiency benefits sought in wearable applications due to complex spline computations and tuning requirements \cite{le2024exploring}, including our own preliminary findings shown in \cref{tab:direct_replace_result}.

This creates a critical dilemma for HAR researchers aiming to leverage cutting-edge architectures: \textbf{How can we harness the potential function approximation power of KANs for complex activity patterns without sacrificing the robustness and computational efficiency essential for practical, real-world wearable HAR systems?} While initial explorations into hybrid KAN models exist \cite{liu2024initial, yang2024kolmogorov, bodner2024convolutional}, there lacks a systematic investigation into how to best integrate KAN components within HAR architectures specifically designed to handle real-world sensor streams effectively and efficiently.

To bridge this gap, we firstly conduct a systematic empirical investigation into strategically integrating KANs and their variants within a strong, efficient, pure MLP-based HAR architecture (MLPHAR \cite{zhou2024mlp}). Our analysis reveals that the placement of KAN modules is critical: they excel at initial data embedding but falter in intermediate feature mixing roles, while specific variants show promise for classification. Based on these insights, we then propose \textbf{KAN-MLP-Mixer}, a novel hybrid architecture that synergistically combines:
\begin{itemize}
    \item[\textbf{1.}] An \textbf{EfficientKAN} module for adaptive data embedding, chosen for its ability to capture input complexities while managing computational overhead.
    \item[\textbf{2.}] Standard \textbf{MLP layers} for intermediate feature mixing, preserving the baseline's proven robustness and efficiency.
    \item[\textbf{3.}] A \textbf{LarctanKAN} module for classification, selected for its smooth, bounded activation potentially offering enhanced stability against noise in the final prediction stage.
\end{itemize}
We evaluate KAN-MLP-Mixer rigorously across eight diverse wearable and ubiquitous HAR datasets. Our results demonstrate that this targeted hybridization achieves a significant average macro F1-score improvement of \textbf{5.33\%} over the robust MLPHAR baseline and substantially outperforms naive full KAN replacement strategies.

The main contributions of this work are:
\begin{itemize}
    \item A \textbf{systematic empirical analysis} of different KAN integration strategies  within an MLP-based HAR framework, identifying performance trade-offs on real-world IMU sensor data. %(full replacement vs. selective placement of various KAN types)
    \item The proposal of \textbf{KAN-MLP-Mixer}, a novel hybrid architecture specifically designed to balance KAN expressiveness with MLP robustness and efficiency for practical HAR.
    \item \textbf{Demonstration of improved performance}, showing that KAN-MLP-Mixer consistently outperforms the MLPHAR baseline across eight diverse public HAR datasets by an average of 5.33\% macro F1-score.
    \item Providing evidence that \textbf{strategic hybridization}, rather than wholesale replacement, offers a viable pathway to leverage advanced architectures like KANs effectively in challenging real-world sensor-based applications common in ubiquitous computing.
    \item \textbf{Extending the hybrid design strategy} across diverse neural backbones, window sizes, and sensing modalities, demonstrating consistent performance improvements under varying conditions through strategic hybridization.
    \item Providing a set of \textbf{practical design guidelines} derived from comprehensive empirical findings to inform the effective application of KANs in sensor-based HAR tasks.
\end{itemize}

While comprehensive benchmarking against the broader HAR state-of-the-art and detailed computational profiling for on-device deployment remain important future steps, our findings provide strong evidence for the potential of carefully constructed KAN-MLP hybrids. This work paves the way for developing more accurate, robust, and ultimately practical HAR systems suitable for the demands of next-generation wearable and mobile technologies.

\section{Related Work}
\label{sec:related_work}

\begin{table}[htbp]
\centering
\small
\caption{Comparative summary of KAN variants}
\begin{tabular}{lllll}
\toprule
\textbf{KAN Type} & \textbf{Basis Function}  & \textbf{Function Form} & \textbf{Major Feature} \\
\midrule
KAN \cite{liu2024kan}& B-spline  & $\phi(x)=w\left(\text{silu}(x) + \sum_i c_i B_i(x)\right)$ & High interpretability \\
EfficientKAN \cite{efficient-kan} & B-spline &  $\phi(x)=w\left(\text{silu}(x) + \sum_i c_i B_i(x)\right)$ & GPU optimized, memory-efficient \\
FastKAN \cite{li2024kolmogorovarnold}& Gaussian RBF &  $\phi(x)=\sum_{i=1}^{K} c_i \exp\left(-\frac{(x - \mu_i)^2}{2\sigma^2}\right)$ & Fast computation \\
WavKAN \cite{bozorgasl2405wav}& Wavelet & $\phi(x)=\sum_{j=0}^{J}\sum_{k=0}^{2^j - 1} w_{j,k}\psi_{j,k}(x)$ & Multiresolution analysis \\
FourierKAN \cite{xu2024FourierKAN}& Sine and Cosine &  $\phi(x)=\sum_{k=1}^{g}(a_k\cos(kx)+b_k\sin(kx))$ & Smooth and differentiable \\
LarctanKAN \cite{chen2024larctan}& Arctan & $\phi(x; k)=\arctan(kx)$ & Compact, efficient activation \\
\midrule
MLP & -- & $\phi(x)=f_{act}(Wx+b)$ & Simple, widely used \\
\bottomrule
\end{tabular}
\label{tab:kan_summary}
\end{table}

This section reviews prior research relevant to our work, focusing on KANs, the role of MLPs in HAR, and the evolution of hybrid architectures in this domain.

\subsection{Kolmogorov-Arnold Networks (KANs): Potential and Challenges}

KANs, inspired by the Kolmogorov–Arnold representation theorem \cite{kolmogorov1957representations, arnold1963, arnold2009functions}, represent a recent shift in neural network design \cite{liu2024kan}. Instead of fixed activations at nodes, KANs employ learnable univariate functions (often splines) on network edges. This grants them significant theoretical expressiveness, allowing them to approximate complex functions accurately, sometimes with fewer parameters than traditional MLPs, particularly on clean, well-structured tasks like symbolic regression or physics modeling \cite{liu2024kan, toscano2025pinns, liu2024kan2, koenig2024kan, genet2024tkan}. This potential for capturing intricate relationships is intriguing for modeling complex human motion patterns from sensor data.

However, the transition from theoretical promise to practical application in ubiquitous computing faces hurdles. KANs' reliance on fine-grained, learnable activation functions makes them susceptible to noise and data irregularities \cite{somvanshi2024survey, shen2025reduced, cang2024can}. This sensitivity is a major concern for wearable sensor HAR, where IMU signals are inherently noisy and variable due to real-world conditions \cite{tseng2023hybrid}. Furthermore, the computational overhead associated with training and evaluating spline-based KANs can be substantial \cite{le2024exploring}, potentially conflicting with the strict resource constraints (power, memory, compute) of wearable devices. Ongoing research aims to mitigate these issues through architectural variants like EfficientKAN \cite{efficient-kan}, FastKAN \cite{li2024kolmogorovarnold}, WavKAN \cite{bozorgasl2405wav}, FourierKAN \cite{xu2024FourierKAN}, and parameter-efficient designs like LarctanKAN \cite{chen2024larctan, chen2024lss}, as well as regularization techniques \cite{altarabichi2024dropkan}. A comparative summary of KAN variants are presented in \cref{tab:kan_summary}. 
An initial study by Liu et al. \cite{liu2024initial} explored KANs for feature extraction in HAR, hinting at their potential but not providing a systematic hybridization strategy for robust, practical use. Thus, while KANs offer a powerful new tool, their direct application to complex, sensor-based HAR scenarios remains challenging.

\subsection{MLPs: The Robust and Efficient Workhorse for HAR}

Despite the advent of newer architectures, MLPs remain a highly relevant and effective tool, particularly in practical HAR applications. Long known as universal function approximators \cite{pinkus1999approximation}, MLPs have seen renewed interest, partly inspired by MLP-Mixer architectures \cite{tolstikhin2021mlp} demonstrating their capability even in complex domains previously dominated by CNNs or Transformers.

Crucially for HAR on wearable devices, MLPs offer a compelling balance of performance, robustness, and \textbf{computational efficiency} \cite{zhou2024mlp, ojiako2023mlps, miyoshi2025applying, liu2021pay}. Their simpler structure with fixed activation functions makes them generally easier to train, more robust to noise, and significantly more lightweight than many alternatives \cite{le2024exploring, fan2021enhanced}. Models like MLPHAR \cite{zhou2024mlp} achieve competitive accuracy on HAR benchmarks using only MLP layers, often with orders of magnitude fewer parameters than CNN or RNN counterparts like DeepConvLSTM \cite{ordonez2016deep} or TinyHAR \cite{zhou2022tinyhar}. This efficiency is vital for deployment on battery-powered wearables. Their established effectiveness and practicality make MLPs not only a strong baseline but also a valuable component for building robust and deployable HAR systems.

\subsection{Hybrid Architectures: Combining Strengths for HAR}

Recognizing that different network components excel at different tasks, hybrid architectures have become common in HAR to leverage complementary strengths. The seminal DeepConvLSTM \cite{ordonez2016deep} combined CNNs for local spatial feature extraction from sensor windows with LSTMs for modeling temporal dependencies, significantly improving performance. This principle has been extended to include attention mechanisms \cite{khan2021attention} and Transformers \cite{dirgova2022wearable, shavit2021boosting,sui2024tramba} combined with CNNs, RNNs, or MLPs \cite{zhou2022tinyhar,zhang2022if,enokibori2024rtsfnet} to capture both local patterns and long-range or global context within activity sequences.

The emergence of KANs opens new possibilities for hybridization. The core motivation is to combine KANs' potential for high-fidelity function approximation with the proven robustness and efficiency of established components like CNNs, RNNs, or, as explored in this work, MLPs \cite{somvanshi2024survey, liu2024initial, yang2024kolmogorov, bodner2024convolutional}. For instance, a KAN module might excel at learning complex initial transformations from raw sensor data, while an MLP or CNN handles subsequent feature aggregation more robustly and efficiently. While the idea of KAN-based HAR hybrids exists, and specific variants like Temporal-KAN have been proposed \cite{somvanshi2024survey}, concrete implementations and systematic studies evaluating how best to integrate KANs into HAR pipelines, particularly considering the noise and efficiency constraints of wearable sensing, are currently lacking.

\section{Empirical Study}

We conducted a comprehensive empirical investigation to explore the impact of integrating various KAN variants into the established \textit{MLPHAR} architecture~\cite{zhou2024mlp}. Specifically, we systematically replaced the original MLP layers in the MLPHAR baseline with different KAN variants, including the original KAN, EfficientKAN, FastKAN, WavKAN, FourierKAN, and LarctanKAN. We evaluated these modifications on eight widely-used HAR datasets.

\subsection{Testbed Model}
% Recent work on KANs has proposed them as promising alternatives to MLPs, particularly for function approximation tasks in deep learning~\cite{liu2024kan}. Unlike conventional neural networks that rely on fixed nonlinearities and scalar weights, KANs employ learnable univariate functions along network edges, offering enhanced expressiveness and a new paradigm for designing neural architectures. Given that MLPs remain foundational components across a wide range of deep learning models, exploring KANs as drop-in replacements for MLP layers represents a compelling research direction.

To systematically evaluate the impact of integrating KANs into existing neural architectures, we selected \textit{MLPHAR}~\cite{zhou2024mlp} as our baseline testbed model. MLPHAR is a purely MLP-based neural architecture specifically tailored for sensor-based HAR tasks.

The MLPHAR model consists of three sequentially concatenated modules: the Data Embedding module, the Feature Mixer module, and the Classifier module. Detailed information about the implementation can be found in the existing work \cite{zhou2024mlp}.
Formally, these modules are arranged in the following order:
\[
\text{Data Embedding} \Rightarrow \text{Feature Mixer} \Rightarrow \text{Classifier}
\]

The straightforward design and robust empirical performance of MLPHAR make it an ideal model to rigorously assess the benefits and limitations of various KAN implementations. By systematically substituting its MLP layers with different KAN variants, we can precisely isolate and quantify the effects of KAN-based components, free from confounding architectural complexities found in models employing convolutional or recurrent modules.

Specifically, MLPHAR was chosen as the test model for several key reasons:

\begin{itemize}
    \item \textbf{Architectural Purity:}
    MLPHAR exclusively employs fully connected layers, intentionally omitting convolutional and recurrent layers commonly found in other HAR architectures. This design choice provides a controlled experimental environment to clearly identify how KAN replacements impact representational capacity and training dynamics compared to MLP.
    
    \item \textbf{Proven Effectiveness:}
    Despite its simplicity, MLPHAR has consistently achieved competitive results across multiple HAR benchmark datasets, as shown in the previous work \cite{zhou2024mlp}, often matching or exceeding the performance of more sophisticated models such as CNNs and LSTMs. Thus, it provides a reliable baseline against which the performance implications of integrating KANs can be accurately measured.
    
    \item \textbf{Practical Efficiency:}
    Due to its compact parameterization and computational efficiency, MLPHAR is particularly well-suited for deployment on edge devices and wearable systems. Evaluating KAN integrations within such an efficiency-critical model directly addresses real-world constraints, ensuring the practical applicability of our findings in ubiquitous computing environments.
\end{itemize}

By employing MLPHAR as our foundational test architecture, this empirical study delivers a clear, consistent, and practically relevant evaluation of various KAN-based architectural designs, providing actionable insights for future development in sensor-based HAR applications.

\subsection{Datasets}
We employed eight widely used benchmark datasets to comprehensively evaluate the impact of integrating various KAN variants into the MLPHAR model. These datasets represent a broad spectrum of HAR scenarios, differing notably in sensor types, body sensor placements, data sampling frequencies, and the complexity of activity classification tasks. 
% In the empirical study, we specially selected one accelerometer sensor mounted in one position from the datasets as shown in \cref{tab:datasets}, and we extended the experiment with multiple sensor modalities and positions in \cref{sec:extening_modality}.
The window size was selected by referring to the existing works \cite{zhou2022tinyhar,zhou2024mlp}.
The raw input data \( X \in \mathbb{R}^{L \times C} \) (with \( L \) as window length and \( C \) as the number of sensor channels) 
is split into \( T \) intervals of length \( \tau \), producing segments \( X_t \in \mathbb{R}^{\tau \times T \times C} \) as the input of the MLPHAR model. 
Detailed characteristics of each dataset are summarized in Table~\ref{tab:datasets}.

By leveraging these diverse datasets, our empirical study robustly evaluates the generalization and performance of various KAN-integrated architectures under realistic and varying conditions characteristic of practical HAR deployments.

\begin{table}[ht]
\centering
\caption{Details of HAR datasets used in experiments.}
\label{tab:datasets}
\begin{threeparttable}
\begin{tabular}{lcccccccc}
\toprule
\textbf{Dataset} & \textbf{Sensors}$^a$ & \textbf{Position}$^b$ & \textbf{Freq. (Hz)} & \textbf{Classes} & \textbf{Channels} & \textbf{Window (s)} & \textbf{T} $^c$& \boldmath$\tau$ $^c$\\
\midrule
\textbf{HAPT} \cite{reyes2013human}        & Acc, Gyro       & Waist       & 50  & 12 & 6 & 2.56 & 8 & 16 \\
\textbf{OPPO} \cite{chavarriaga2013opportunity}        & Acc, Gyro, Mag  & Lower Arm   & 30  & 18 & 9 & 1.00 & 3 & 10 \\
\textbf{DG} \cite{bachlin2009wearable}          & Acc             & Leg         & 64  & 2  & 3 & 1.00 & 4 & 16 \\
\textbf{PAMAP2}\cite{reiss2012introducing}       & Acc, Gyro       & Right Hand  & 33  & 12 & 6 & 3.00 & 9 & 11 \\
\textbf{Skodar} \cite{zappi2008activity}      & Acc             & Right Wrist & 30  & 10 & 3 & 3.00 & 5 & 18 \\
\textbf{DSADS} \cite{zhang2015recognizing}       & Acc, Gyro, Mag  & Arm         & 25  & 19 & 9 & 5.00 & 5 & 25 \\
\textbf{MotionSense}\cite{malekzadeh2021dana}  & Acc, Gyro       & Waist       & 50  & 6  & 6 & 2.56 & 8 & 16 \\
\textbf{MHEALTH}\cite{banos2014MHEALTHdroid}      & Acc, Gyro       & Arm         & 50  & 13 & 6 & 2.56 & 8 & 16 \\
\bottomrule
\end{tabular}
\begin{tablenotes}
\item[a] only the IMU sensor is selected
\item[b] only sensor in one position is selected
\item[c] The raw input data \( X \in \mathbb{R}^{L \times C} \) (with \( L \) as window length and \( C \) as the number of sensor channels) 
is split into \( T \) intervals of length \( \tau \), producing segments \( X_t \in \mathbb{R}^{\tau \times T \times C} \) as the input of the MLPHAR Model. 
\end{tablenotes}
\end{threeparttable}
\end{table}

\subsection{Experiment Setup}
%GPU Nvidia A6000
%epoch 200
%early stop 7
%learning rate 0.001
%optimiser: Adam
%most of datasets are using leave one subjects out cross validation, except MotionSensen (leave group out) and Skodar using leave one session out.
%The macro F1 score was selected as the metric 

All experiments were conducted on an NVIDIA A6000 GPU. We trained each model for up to 200 epochs, employing early stopping with a patience of 7 epochs to prevent overfitting. Model parameters were optimized using the Adam optimizer with an initial learning rate of 0.001.
We adopted subject-independent evaluation strategies appropriate for each dataset to ensure fair and realistic assessments. Specifically, most datasets utilized a Leave-One-Subject-Out (LOSO) cross-validation scheme, where data from each subject was iteratively reserved as the test set while data from all remaining subjects served as the validation and training set. Two exceptions were the MotionSense dataset, which employed Leave-Group-Out validation, and the Skodar dataset, evaluated using Leave-One-Session-Out cross-validation.
Model performance was assessed using the macro F1-score, providing a balanced metric suitable for handling class-imbalanced HAR datasets and effectively capturing performance across all activity categories.
To ensure reproducibility and consistent results, the initial parameters were generated with five seeds (1-5), and the average was reported.
In the empirical experiment, only the data from the accelerometer in a single body part was used; we further extended the experiment to multiple sensors in multiple positions in \cref{sec:extening_modality}.

This empirical study consists of two sequential parts, each designed to analyze the integration of KANs into the MLPHAR architecture from a different perspective.
In the first part, we replace all MLP (or linear) layers in the original MLPHAR model with KAN or one of its variants to construct fully KAN-based models. This setup allows us to assess the performance and limitations of pure KAN architectures in the context of HAR tasks, and to compare them directly against the original MLP-based baseline.
In the second part, we investigate the effects of partial KAN integration by selectively inserting KAN components at different positions within the MLPHAR architecture, specifically, the data embedding, feature mixer, or classifier modules. 
This staged integration enables us to precisely isolate and quantify the individual contributions of KAN modules at each stage of the network, offering deeper insights into where KANs are most effective within a modern MLP-based HAR pipeline.

To ensure fair comparison, we maintain consistent input/output dimensions and layer counts between MLP and KAN configurations, with the data embedding module set to a hidden dimension of 16 across all datasets. 
KAN configuration details are as follows: KAN and EfficientKAN use a grid size of 5, spline degree 3, with grid range $[-1, 1]$; FastKAN uses grid size 8, grid range $[-2, 2]$; FourierKAN uses grid size 5, and the Mexican hat function was selected in WavKAN.
Further analysis on the effects of hidden dimension (in MLPHAR) and grid size (in EfficientKAN) is discussed in \cref{sec: evaluate_efficiency}.

% \subsection{Investigating KANs for HAR}

\subsection{Direct Replacement of all MLPs with KANs}

\begin{table}[ht]
\scriptsize
\centering
\caption{Result Summary of direct replacement of MLPs with KANs in the MLPHAR  baseline \cite{zhou2024mlp} across eight HAR datasets. 
Each cell shows the macro F1 score on a specific dataset.
The highest mean macro F1 score per dataset is highlighted in \textcolor{Gold}{\textbf{Gold}}, the second-best in \textcolor{Silver}{\textbf{Silver}} and the third-best in \textcolor{Bronze}{\textbf{Bronze}}. 
The last column reports the average relative improvement (\%) of each model across all datasets over the baseline. (macro-F1 ± std).
% Direct replacement of MLP layers with KANs generally leads to performance degradation, except for marginal gains in isolated cases (e.g., \texttt{dg} and \texttt{oppo}).
}

\begin{tabular}{cccccccccc}
\toprule
\textbf{Model} & \textbf{DG} & \textbf{DSADS} & \textbf{PAMAP2} & \textbf{OPPO} & \textbf{Skodar} & \textbf{HAPT} & \textbf{MotionSense} & \textbf{MHEALTH} & \textbf{Gain (\%)} \\
\midrule
KAN & \textcolor{Silver}{\textbf{0.591±0.164}} & 0.434±0.228 & 0.322±0.304 & 0.318±0.198 & \textcolor{Bronze}{\textbf{0.396±0.431}} & 0.297±0.339 & 0.420±0.387 & 0.306±0.373& -36.66 \\
EfficientKAN & \textcolor{Gold}{\textbf{0.592±0.164}} & 0.432±0.228 & 0.322±0.304 & 0.317±0.197 & 0.395±0.430 & 0.298±0.340 & 0.421±0.388 & 0.305±0.372& -36.74 \\
FastKAN & 0.543±0.116 & \textcolor{Silver}{\textbf{0.503±0.078}} & \textcolor{Silver}{\textbf{0.474±0.200}} & \textcolor{Silver}{\textbf{0.415±0.124}} & \textcolor{Silver}{\textbf{0.478±0.190}} & \textcolor{Silver}{\textbf{0.604±0.126}} & \textcolor{Silver}{\textbf{0.753±0.025}} & \textcolor{Silver}{\textbf{0.627±0.161}}& -12.37 \\
WavKAN & 0.533±0.040 & 0.377±0.074 & 0.435±0.151 & \textcolor{Gold}{\textbf{0.420±0.086}} & 0.087±0.013 & 0.480±0.037 & 0.654±0.051 & 0.495±0.049& -28.20 \\
FourierKAN & 0.570±0.218 & 0.015±0.007 & 0.037±0.061 & 0.114±0.010 & 0.051±0.010 & 0.263±0.206 & 0.666±0.111 & 0.064±0.066& -66.45 \\
LarctanKAN & \textcolor{Bronze}{\textbf{0.583±0.187}} & \textcolor{Bronze}{\textbf{0.480±0.105}} & \textcolor{Bronze}{\textbf{0.440±0.269}} & 0.185±0.151 & 0.053±0.156 & \textcolor{Bronze}{\textbf{0.547±0.267}} & \textcolor{Bronze}{\textbf{0.672±0.310}} & \textcolor{Bronze}{\textbf{0.592±0.310}}& -28.99 \\
Baseline (MLP) & 0.580±0.101 & \textcolor{Gold}{\textbf{0.504±0.090}} & \textcolor{Gold}{\textbf{0.540±0.186}} & \textcolor{Bronze}{\textbf{0.411±0.175}} & \textcolor{Gold}{\textbf{0.844±0.017}} & \textcolor{Gold}{\textbf{0.685±0.046}} & \textcolor{Gold}{\textbf{0.836±0.026}} & \textcolor{Gold}{\textbf{0.747±0.114}}& 0.00 \\
\hline
\end{tabular}

\label{tab:direct_replace_result}
\end{table}

To evaluate the direct effectiveness of KANs in HAR tasks, we replaced all MLP (or linear) layers in the MLPHAR baseline model with KAN-based alternatives. The results are summarized in \cref{tab:direct_replace_result}, which reports the average macro F1 score across eight benchmark HAR datasets.
As shown in the  in \cref{tab:direct_replace_result}, the baseline MLP model (MLPHAR) consistently outperforms all KAN-based variants on the majority of datasets. Specifically, MLP achieves the best results on seven out of eight datasets, with substantial margins in several cases such as \texttt{PAMAP2}, \texttt{MotionSense}, and \texttt{MHEALTH}. The only exceptions are found in the \texttt{DG} dataset, where the vanilla \texttt{KAN} slightly exceeds the baseline (0.591 vs. 0.580), and in \texttt{OPPO}, where \texttt{WavKAN} shows a marginal gain (0.420 vs. 0.411). These two cases, however, are isolated and do not indicate a general trend.

Among the KAN variants tested, \texttt{FastKAN} and \texttt{LarctanKAN} demonstrate relatively competitive performance. \texttt{FastKAN}, in particular, achieves the smallest drop in average accuracy (only --12.37\% relative to the MLP baseline), suggesting its potential when adapted properly. In contrast, \texttt{FourierKAN} performs significantly worse across all datasets, with the steepest performance degradation (--66.45\%), raising questions about the suitability of Fourier basis representations for HAR tasks in this form.

The consistent performance gap observed between the MLP baseline and KAN-based replacements suggests that direct substitution is not effective. This may be attributed to several factors. First, KANs introduce a different representational inductive bias compared to standard MLPs, which could disrupt feature hierarchies learned through dense layers. Second, KANs rely on spline-based function approximations that may require task-specific tuning of hyperparameters, such as the number of grids or the interpolation strategy. Lastly, their optimization behavior and sensitivity to initialization could lead to suboptimal convergence when used without tailored training routines. 
% (The poor information could be that during the training stage the variables can shift out of the domain, so an additional rescaling of the spline grids are introduced.)

% Despite the overall negative trend, certain KAN variants showed promise on specific datasets. This opens opportunities for more nuanced use of KANs in HAR systems. Rather than replacing all dense layers, future work could explore hybrid architectures where KANs are integrated only at select layers. Additionally, dedicated tuning of KAN-specific components—such as regularization schemes, learning rates, or activation smoothing—might bridge the performance gap.

In a short conclusion, while the theoretical flexibility of KANs makes them appealing, our results show that a naïve replacement of MLPs in HAR pipelines is insufficient. More targeted design and careful tuning are required to unlock the full potential of KANs in this domain.

\subsection{Performance of Selective KAN Integration}
\cref{tab:result_summary_position} summarizes the experimental outcomes of these configurations across eight benchmark datasets. The primary observations drawn from these results are:

\begin{itemize}
    \item \textbf{KAN in Data Embedding (K-M-M)}: 
    Integrating KAN variants at the Data Embedding module typically yielded positive results. Notably, standard KAN and EfficientKAN in the embedding position resulted in average performance improvements of approximately 3.38\% and 3.15\%, respectively. This indicates that KAN modules effectively capture initial nonlinear relationships and representations from raw sensor data.

    \item \textbf{KAN in Feature Mixer (M-K-M)}:
    Replacing the Feature Mixer module with KAN variants generally resulted in substantial performance degradation (ranging approximately from -7\% to -25\%). Such consistent performance drops suggest that KAN modules might not be well-suited for the role of intermediate feature transformation, possibly due to difficulties handling feature interactions at deeper network layers.

    \item \textbf{KAN in Classifier (M-M-K)}:
    The integration of KANs at the Classifier module produced mixed results. While most variants showed neutral or negative impacts, LarctanKAN showed slight positive effects (+1.36\% improvement), suggesting that certain KAN designs could offer advantages in classification scenarios, potentially due to their smooth activation functions aiding decision boundary formation. This result aligns with the existing work \cite{chen2024larctan}.
\end{itemize}

Overall, these selective integration results demonstrate the importance of module-specific placements when employing KAN-based architectures in HAR. Optimal performance gains from KANs are achieved by carefully identifying the modules within which their unique functional approximation capabilities can be most effectively leveraged.

\section{Proposed Hybrid Architecture}

\begin{figure*}[hbt]
\centering
\includegraphics[width=0.99\linewidth]{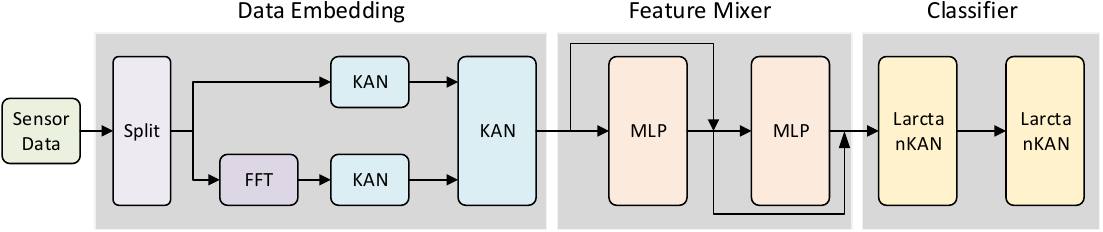}
\caption{The proposed hybrid network architecture KAN-MLP-Mixer based on an empirical study. It consists of three modules: KAN-based data embedding, MLP-based feature mixer and LarctanKAN-based classifier. This proposed hybrid network is built based on the MLPHAR framework \cite{zhou2024mlp}. During implementation, we employed \textbf{EfficientKAN} in place of the original KAN, as it delivers nearly identical performance while offering significantly improved memory efficiency and training speed. (\textit{Split}: The raw input data \( X \in \mathbb{R}^{L \times C} \) (with \( L \) as window length and \( C \) as the number of sensor channels) is split into \( T \) intervals of length \( \tau \), producing segments \( X_t \in \mathbb{R}^{\tau \times T \times C} \) as the input of the neural network. \textit{FFT}: Fast Fourier Transform)}
\label{fig:kan-mlp-mixer}
\end{figure*}

Based on the findings of our empirical study of KAN integration, we propose a new hybrid architecture, \textbf{KAN-MLP-Mixer},  strategically combining the strengths of both KAN and MLPs to improve the performance in HAR tasks.

\subsection{Design Rationale}

Our experimental results reveal that KANs perform best when applied to the input embedding stage, where they capture complex nonlinear relationships from raw sensor data. In contrast, substituting KANs into the feature mixer significantly degrades performance, likely due to optimization difficulties and overfitting. Finally, LarctanKAN demonstrates a unique advantage when used in the classifier, offering improved generalization and smooth decision boundaries.

Based on these insights, we adopt a modular hybrid architecture as shown in \cref{fig:kan-mlp-mixer}.
This design leverages the expressiveness of KANs at the input level, preserves the stability and scalability of MLPs for feature transformation, and enhances classification robustness via LarctanKAN’s bounded, smooth activation function.

\subsection{Architecture Details}

\begin{itemize}
    \item \textbf{Data Embedding (EfficientKAN):}
    The raw input signal is passed through a KAN-based embedding module. We adopt the EfficientKAN variant due to its lower memory cost and strong performance in the data embedding role, as observed in our experiments. This module transforms raw sensor input into a high-level feature representation using spline-based function learning.

    \item \textbf{Feature Mixer (MLP):}
    The transformed features are then passed through a standard MLP block, which remains unchanged from the original MLPHAR. This module performs the core latent feature transformation and mixing. Our results show that MLPs outperform all tested KANs in this role, providing better generalization and stability.

    \item \textbf{Classifier (LarctanKAN):}
    The final layer is a LarctanKAN classifier, which replaces the original MLP-based classification head. The arctangent-based activation offers smooth gradients and bounded outputs, which improve robustness and classification performance.
\end{itemize}

\subsection{Expected Benefits}

This hybrid architecture is designed to: (1)\textbf{Enhance expressiveness} in early feature extraction using KANs. (2)\textbf{Retain robustness and efficiency} in mid-layer processing via MLPs. (3)\textbf{Improve generalization and stability} in classification using LarctanKAN.
% \begin{itemize}
%     \item \textbf{Enhance expressiveness} in early feature extraction using KANs.
%     \item \textbf{Retain robustness and efficiency} in mid-layer processing via MLPs.
%     \item \textbf{Improve generalization and stability} in classification using LarctanKAN.
% \end{itemize}

In doing so, it addresses the performance degradation seen in fully KAN-based models while still leveraging their theoretical strengths in function approximation. We demonstrate in the following section that this architecture consistently outperforms both pure KAN and pure MLP baselines across multiple HAR benchmarks.

\section{Evaluation of the Proposed Hybrid Architecture}

To comprehensively evaluate our proposed hybrid architecture (\textbf{KAN-MLP-Mixer}), we benchmarked its performance across eight diverse and representative HAR datasets, comparing against both the original MLPHAR baseline and previously tested selective KAN configurations. \cref{tab:result_summary_position} summarizes the macro F1 scores obtained, clearly highlighting our proposed model's effectiveness.

\subsection{Overall Performance}
Our proposed KAN-MLP-Mixer architecture consistently outperforms all previously evaluated configurations, achieving an average macro F1 improvement of \textbf{+5.33\%} relative to the original MLPHAR baseline. Such significant gains underscore the complementary advantages of strategically integrating KAN modules into pure MLP-based models, confirming our initial hypothesis that careful placement of KANs can substantially enhance HAR performance.

% \subsection{Dataset-wise Analysis and Observations}
Our hybrid model consistently outperformed baseline architectures, demonstrating substantial improvements in datasets such as OPPO (+5.90\%), Skodar (+4.38\%), MHEALTH (+5.52\%), DSADS (+3.10\%), and HAPT (+2.00\%), showcasing the hybrid approach’s superior ability to handle modality fusion, noise resilience, and fine-grained activity distinctions. Even in simpler or structured scenarios (DG, MotionSense, PAMAP2), the hybrid design achieved notable improvements by strategically leveraging EfficientKAN’s expressive embedding and LarctanKAN’s robust classification capabilities, balanced by the generalization strength of MLP-based mixers. These results validate our selective integration of KAN modules, confirming their practical effectiveness and adaptability across diverse real-world wearable HAR contexts.

\subsection{Comparative Insights and Module Contributions}
Our results yield several key insights regarding module-specific contributions within the KAN-MLP-Mixer architecture:

\begin{itemize}
    \item \textbf{KAN-based Data Embedding (EfficientKAN):} Consistently improves early-stage feature extraction, particularly effective with noisy and raw sensor signals, confirming prior observations that expressive embedding functions significantly boost performance in HAR tasks.

    \item \textbf{MLP-based Feature Mixer:} Critical for maintaining high generalization performance. Our empirical results reinforce earlier findings that MLP layers remain highly effective at deeper network levels, especially when complex feature interactions and stability are required.

    \item \textbf{LarctanKAN Classifier:} Provides substantial advantages in classification accuracy, leveraging smooth and bounded activation functions that create well-defined and stable decision boundaries, crucial in noisy HAR data contexts.
\end{itemize}

This nuanced combination leverages each module’s specific strengths, resulting in the superior overall performance observed. 

% \subsection{Generalization and Stability}
% Compared to previous approaches that demonstrated substantial variability across datasets, our proposed hybrid model exhibits remarkable stability. Its consistently high-ranking performance across multiple sensor placements, and classification complexities demonstrates a generalized efficacy crucial for ubiquitous computing and wearable applications.

\subsection{Generalization and Practical Implications}
% In summary, the proposed KAN-MLP-Mixer hybrid architecture demonstrates the clear benefits of strategic KAN integration into MLP-based HAR models. The achieved improvements in both accuracy and robustness across diverse datasets underscore its practical applicability, making this architecture a promising candidate for deployment in real-world HAR applications on mobile and wearable platforms. These empirical findings provide actionable insights for researchers and practitioners aiming to design robust, accurate, and efficient activity recognition systems.

Compared to previous approaches that exhibited significant performance variability across datasets, the proposed KAN-MLP-Mixer hybrid architecture demonstrates remarkable stability and generalization. Its consistently superior performance across diverse sensor placements and varying classification complexities underscores its suitability for real-world wearable and ubiquitous computing applications. These empirical findings highlight the clear advantages of strategically integrating KAN modules into robust MLP-based HAR frameworks, offering actionable guidance for researchers and practitioners seeking to build accurate, efficient, and reliable HAR systems on mobile and wearable devices.

\begin{table}[ht]
\scriptsize
\centering
\caption{Comparison of Model Performance across Various Datasets. This table provides the mean F1 score for different models evaluated on various datasets. The top three models for each dataset are highlighted: first place in \textcolor{Gold}{\textbf{Gold}}, second place in \textcolor{Silver}{Silver}, and third place in \textcolor{Bronze}{\textbf{Bronze}}.(macro-F1 ± std).}
\label{tab:result_summary_position}
\begin{tabular}{cccccccccc}
\toprule
\textbf{Model} & \textbf{DG} & \textbf{DSADS} & \textbf{PAMAP2} & \textbf{OPPO} & \textbf{Skodar} & \textbf{HAPT} & \textbf{MotionSense} & \textbf{MHEALTH} & \textbf{Gain (\%)} \\
\midrule
\multicolumn{9}{c}{KAN} \\
\hline
K-M-M & \textcolor{Silver}{\textbf{0.589±0.093}} & 0.509±0.071 & 0.549±0.192 & \textcolor{Gold}{\textbf{0.478±0.090}} & \textcolor{Bronze}{\textbf{0.860±0.020}} & 0.697±0.040 & 0.831±0.024 & \textcolor{Bronze}{\textbf{0.772±0.095}}& 3.38 \\
M-K-M & 0.586±0.175 & 0.416±0.219 & 0.450±0.274 & 0.256±0.187 & 0.525±0.419 & 0.487±0.323 & 0.599±0.357 & 0.457±0.378& -25.62 \\
M-M-K & 0.571±0.080 & 0.517±0.091 & 0.549±0.183 & 0.412±0.155 & 0.834±0.022 & 0.697±0.043 & 0.829±0.027 & 0.755±0.114& 0.46 \\
\hline
\multicolumn{9}{c}{EfficientKAN} \\
\hline
K-M-M & \textcolor{Bronze}{\textbf{0.587±0.094}} & 0.507±0.071 & 0.547±0.192 & \textcolor{Silver}{\textbf{0.477±0.086}} & 0.858±0.017 & \textcolor{Bronze}{\textbf{0.698±0.041}} & 0.831±0.025 & 0.769±0.097& 3.15 \\
M-K-M & 0.584±0.175 & 0.419±0.219 & 0.449±0.274 & 0.252±0.184 & 0.526±0.420 & 0.486±0.322 & 0.598±0.356 & 0.457±0.378& -25.71 \\
M-M-K & 0.572±0.081 & 0.517±0.089 & 0.545±0.183 & 0.411±0.156 & 0.835±0.016 & 0.696±0.044 & 0.829±0.025 & 0.757±0.117& 0.39 \\
\hline
\multicolumn{9}{c}{FastKAN} \\
\hline
K-M-M & 0.562±0.080 & 0.484±0.178 & 0.482±0.242 & 0.388±0.155 & 0.426±0.337 & 0.672±0.044 & 0.791±0.028 & 0.739±0.091& -10.16 \\
M-K-M & 0.558±0.081 & 0.451±0.067 & 0.512±0.175 & 0.434±0.080 & 0.808±0.028 & 0.657±0.098 & 0.803±0.022 & 0.721±0.102& -3.72 \\
M-M-K & 0.560±0.080 & 0.486±0.075 & 0.531±0.184 & 0.335±0.181 & \textcolor{Gold}{\textbf{0.887±0.018}} & 0.673±0.101 & 0.828±0.021 & 0.746±0.109& -3.14 \\
\hline
\multicolumn{9}{c}{WavKAN} \\
\hline
K-M-M & 0.553±0.066 & \textcolor{Bronze}{\textbf{0.523±0.078}} & \textcolor{Gold}{\textbf{0.563±0.188}} & 0.464±0.081 & 0.793±0.026 & 0.610±0.060 & 0.787±0.029 & 0.735±0.107& -1.02 \\
M-K-M & 0.523±0.052 & 0.429±0.081 & 0.452±0.158 & 0.398±0.054 & 0.729±0.041 & 0.562±0.051 & 0.721±0.027 & 0.575±0.080& -14.07 \\
M-M-K & 0.548±0.079 & 0.363±0.093 & 0.467±0.175 & 0.353±0.089 & 0.638±0.128 & 0.609±0.051 & 0.765±0.053 & 0.622±0.086& -15.22 \\
\hline
\multicolumn{9}{c}{FourierKAN} \\
\hline
K-M-M & 0.543±0.050 & 0.311±0.180 & 0.518±0.176 & 0.377±0.147 & 0.714±0.024 & 0.641±0.043 & 0.712±0.125 & 0.667±0.091& -13.02 \\
M-K-M & 0.541±0.072 & 0.438±0.077 & 0.496±0.160 & 0.431±0.084 & 0.721±0.040 & 0.643±0.043 & 0.777±0.021 & 0.666±0.094& -7.70 \\
M-M-K & 0.547±0.059 & 0.385±0.073 & 0.458±0.155 & 0.406±0.060 & 0.528±0.052 & 0.615±0.036 & 0.783±0.019 & 0.586±0.067& -15.18 \\
\hline
\multicolumn{9}{c}{LarctanKAN} \\
\hline
K-M-M & 0.568±0.080 & 0.507±0.084 & 0.552±0.181 & 0.408±0.175 & 0.810±0.022 & 0.682±0.040 & 0.830±0.027 & 0.755±0.098& -0.54 \\
M-K-M & 0.579±0.176 & \textcolor{Gold}{\textbf{0.543±0.095}} & 0.440±0.271 & 0.367±0.206 & 0.426±0.397 & 0.563±0.275 & 0.724±0.265 & 0.613±0.288& -15.07 \\
M-M-K & 0.584±0.101 & 0.513±0.078 & \textcolor{Bronze}{\textbf{0.559±0.187}} & 0.402±0.168 & 0.834±0.027 & \textcolor{Silver}{\textbf{0.698±0.037}} & \textcolor{Silver}{\textbf{0.837±0.026}} & \textcolor{Silver}{\textbf{0.793±0.109}}& 1.36 \\
\hline
MLPHAR & 0.580±0.101 & 0.504±0.090 & 0.540±0.186 & 0.411±0.175 & 0.844±0.017 & 0.685±0.046 & \textcolor{Bronze}{\textbf{0.836±0.026}} & 0.747±0.114& 0.00 \\
\hline

\hline
KAN-MLP-Mixer & \textcolor{Gold}{\textbf{0.598±0.098}} & \textcolor{Silver}{\textbf{0.535±0.081}} & \textcolor{Silver}{\textbf{0.560±0.192}} & \textcolor{Bronze}{\textbf{0.470±0.080}} & \textcolor{Silver}{\textbf{0.881±0.019}} & \textcolor{Gold}{\textbf{0.705±0.040}} & \textcolor{Gold}{\textbf{0.840±0.026}} & \textcolor{Gold}{\textbf{0.802±0.102}}& 5.33 \\
\hline
\end{tabular}

\end{table}

\section{Ablation Study: Module-Level Analysis of Hybrid Design}

\begin{figure*}
    \centering
      \begin{subfigure}[b]{0.33\textwidth}
     
      \includegraphics[width=\linewidth]{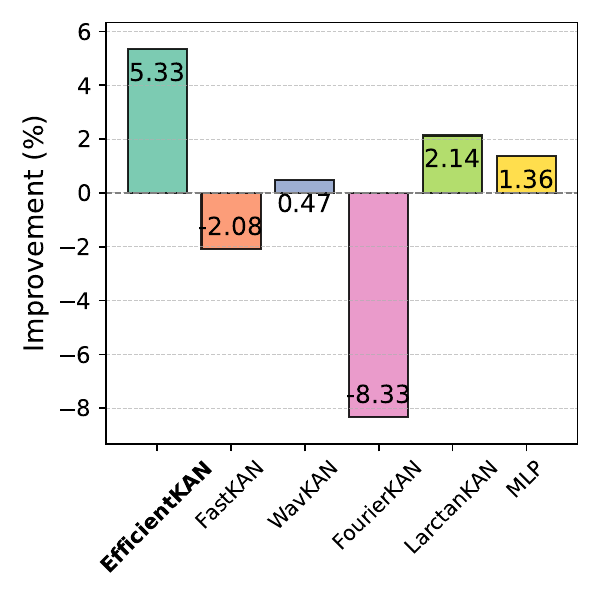}
      \caption{
        Data Embedding
      }
      \label{fig:embedding-ablation}
    \end{subfigure}
    \hfill
      \begin{subfigure}[b]{0.33\textwidth}
      \includegraphics[width=\linewidth]{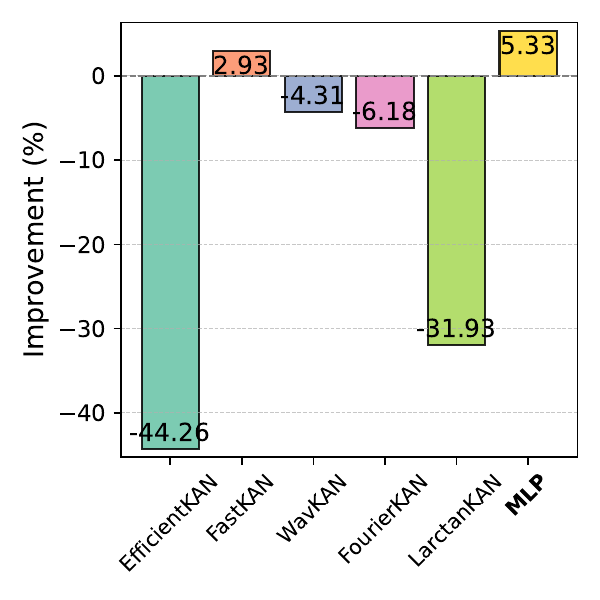}
      \caption{
        Feature Mixer
      }
      \label{fig:mixer-ablation}
    \end{subfigure}
    \hfill
      \begin{subfigure}[b]{0.33\textwidth}
      \includegraphics[width=\linewidth]{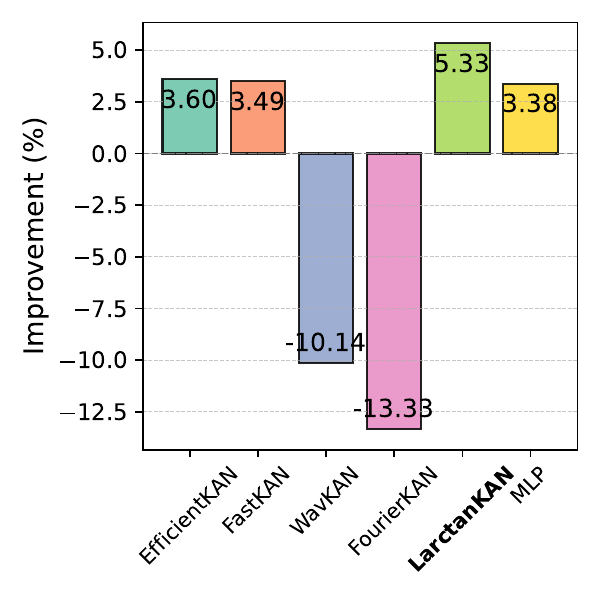}
      \caption{
        Classifier
      }
      \label{fig:classifier-ablation}
    \end{subfigure}
    \caption{Average performance improvement compared to the MLPHAR baseline across eight datasets using the hybrid model with different \textbf{(a)} data embedding, \textbf{(b)} feature mixer, and \textbf{(c)} classifier modules. 
    In (a), the feature mixer is fixed as MLP and the classifier as LarctanKAN. 
    In (b), the data embedding is fixed as EfficientKAN and the classifier as LarctanKAN. 
    In (c), the data embedding is EfficientKAN and the feature mixer is MLP.}
    \label{fig:ablation-all}
\end{figure*}

To systematically examine the role and effectiveness of each component in the hybrid architecture inspired by the empirical study, we conducted a comprehensive ablation study across three core modules: the data embedding layer, feature mixer, and classifier. For each module, we tested six alternatives, five KAN variants (EfficientKAN, FastKAN, WavKAN, FourierKAN, LarctanKAN) and a standard MLP, while keeping the other two components fixed. The results, shown in \cref{fig:ablation-all}, report average performance improvements compared to the original MLPHAR model across eight HAR datasets.

In \cref{fig:embedding-ablation}, EfficientKAN achieves the highest improvement (+5.33\%), validating its expressiveness and efficiency for modeling raw, smooth IMU signals. LarctanKAN (+2.14\%) and MLP (+1.36\%) also deliver modest improvements, while WavKAN (+0.47\%) shows limited gains. In contrast, FastKAN (-2.08\%) and FourierKAN (-8.33\%) underperform significantly, indicating that not all spline-based or basis-function approaches are suitable for early-stage processing in noisy or variable sensor conditions.

\cref{fig:mixer-ablation} highlights the importance of choosing a stable and generalizable mixer module. When EfficientKAN and LarctanKAN are fixed, MLP provides the best improvement (+5.33\%), reaffirming its effectiveness in latent feature transformation. FastKAN also performs well (+2.93\%), but other KAN variants, particularly EfficientKAN (-44.26\%), FourierKAN (-6.18\%) and LarctanKAN (-31.93\%), degrade performance sharply. This suggests that KANs are more prone to instability and overfitting in deeper latent spaces, likely due to sensitivity to input distributions and higher-order feature interactions.

As shown in \cref{fig:classifier-ablation}, LarctanKAN outperforms all other classifier options (+5.33\%), confirming its ability to model complex decision boundaries without instability. MLP (+3.38\%), EfficientKAN (+3.60\%), and FastKAN (+3.49\%) provide moderate improvements. However, WavKAN (-10.14\%) and FourierKAN (-13.33\%) again show severe performance drops, likely due to poorly shaped output manifolds or unbounded activation behaviors.

These results strongly reinforce the modular strategy behind our final model:
\begin{itemize}
    \item \textbf{EfficientKAN is most suitable as a data embedding layer}, effectively extracting rich features from smooth IMU signals.
    \item \textbf{MLP remains the most stable and generalizable feature mixer}, ensuring robustness in high-dimensional latent spaces.
    \item \textbf{LarctanKAN provides the most effective classifier module}, offering smooth yet discriminative decision boundaries critical for HAR classification.
\end{itemize}

In addition, this study also highlights that KAN variants are not interchangeable; their performance is highly sensitive to position within the network. When misapplied, they can cause significant degradation, as seen with FourierKAN in both embedding and classifier roles, and EfficientKAN in the feature mixer. 

Thus, this ablation study provides both quantitative and architectural support for the hybrid design: EfficientKAN–MLP–LarctanKAN. 
% Each module is placed where it performs best, resulting in a model that is both interpretable and performant for ubiquitous HAR scenarios.

\section{Discussion}
In this section, we first evaluate the generalizability of the proposed hybrid design across multiple sensing modalities, diverse window sizes, and a variety of neural backbones. We then evaluate parameter and computational efficiency of hybrid KAN designs. Finally, we discuss the limitations of the current work, outline potential directions for future research, and offer practical design guidelines for applying KANs in sensor-based HAR systems.

\subsection{Extending Hybrid Design Across Multiple Modalities}
\label{sec:extening_modality}
To evaluate the generalizability of the KAN-MLP-Mixer hybrid architecture across different sensing modalities and channels, we conducted experiments under three sensor configurations as shown in \cref{tab:sensor-conf}: (1) single 3-axis accelerometer (ACC) in single body position, (2) single IMU, and (3) multiple sensors. 
We exclude the dataset Skodar, DG and HAPT, because only acceleration sensor data is available in the first two datasets, and only 6-channel IMU data is included in the HAPT dataset.

% \begin{table}[ht]
% \centering
% \caption{Sensor channel configuration and the performance improvement by KAN-MLP-Mixer (The gain is the average macro F1 score improvement across the five dataset by hybrid KAN-MLP-Mixer compared to pure MLPHAR model)}
% \label{tab:sensor-conf}
% % \scriptsize
% \begin{threeparttable}
    
% \begin{tabular}{lcccccc}
% \toprule
% \textbf{Configuration} & \textbf{DSADS} & \textbf{PAMAP2} & \textbf{OPPO} & \textbf{MotionSense} & \textbf{MHEALTH} & \textbf{ Average gain (\%)} \\
% \midrule
% Single acceleration & 3 & 3 & 3   & 3  & 3 & 6.43\\
% Single IMU $^1$ & 9 & 6 & 9        & 6 & 6  & 4.68\\
% Multi-sensors$^2$ & 45 & 18 & 77  & 12 & 12 & 2.17\\
% \bottomrule
% \end{tabular}
% \begin{tablenotes}
% \item[1] 6-channel configuration includes 3-axis accelerometer and 3-axis gyroscope. another 3-axis magnetic information is included in a 9-channel configuration.
% \item[2] Multiple sensor configuration includes multiple IMUs on different body parts and a single IMU with additional feature channels like rotation angle.
% \end{tablenotes}
% \end{threeparttable}
% \end{table}

\begin{table}[ht]
\centering
\caption{Sensor channel configuration and the performance improvement by KAN-MLP-Mixer (The gain is the average macro F1 score improvement across the five dataset by hybrid KAN-MLP-Mixer compared to pure MLPHAR model)}
\label{tab:sensor-conf}
% \scriptsize
\begin{threeparttable}
    
\begin{tabular}{l|c|c|c}
\toprule
\multirow{2}{*}{\textbf{Dataset}} & \textbf{ \multirow{2}{*}{Channels for One Acc}} & \multirow{2}{*}{\textbf{Channels for Single IMU $^1$}} & \multirow{2}{*}{\textbf{Channels for Multiple Sensors$^2$}} \\
 &  &  &   \\
\midrule
 \textbf{DSADS} & 3 & 9 & 45  \\
 \textbf{PAMAP2} & 3 & 6 & 18  \\
  \textbf{OPPO} & 3 & 9 & 77  \\
  \textbf{MotionSense} & 3 & 6 & 12  \\
  \textbf{MHEALTH} & 3 & 6 & 12  \\
\bottomrule
\bottomrule

\textbf{Macro F1 Gain (\%)} & 6.43 & 4.68 & 2.17 \\
\bottomrule
\end{tabular}
\begin{tablenotes}
\item[1] 6-channel configuration includes 3-axis accelerometer and 3-axis gyroscope. another 3-axis magnetic information is included in a 9-channel configuration.
\item[2] Multiple sensor configuration includes multiple IMUs on different body parts and a single IMU with additional feature channels like rotation angle.
\end{tablenotes}
\end{threeparttable}
\end{table}

\begin{figure*}[hbt]
\centering
\includegraphics[width=\linewidth]{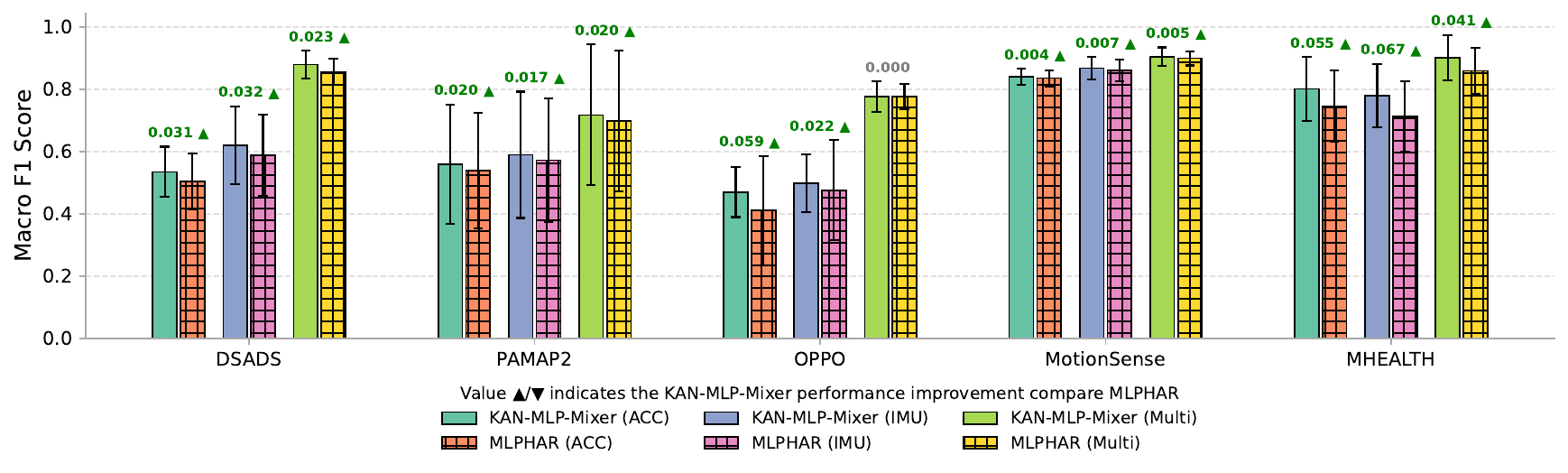}
\caption{Performance comparison for KAN-MLP-Mixer and MLPHAR models on five datasets under three sensor configurations (single ACC, single IMU, Multiple sensors). Numerical annotations show the absolute performance difference between models for each condition; $\blacktriangle$ favors KAN-MLP-Mixer, $\blacktriangledown$ favors MLPHAR.}
\label{fig:result_cross_modality}
\end{figure*}

 As shown in \cref{fig:result_cross_modality} and \cref{tab:sensor-conf}, the hybrid models KAN-MLP-Mixer consistently outperformed the MLPHAR baseline across all five public datasets. The KAN-MLP-Mixer model achieved an average macro F1 score improvement of 6.43\% with a single 3-channel accelerometer input, 4.68 \% with a single IMU sensor input, and 2.17\% with multiple channel sensors, compared to the MLPHAR baseline. These results demonstrate that the benefits of introducing KAN-based feature embeddings and the LarctanKAN module are not confined to a specific modality or sensor richness level.
 The largest performance gains were observed under the single-accelerometer setting, where sensor information is relatively sparse. This finding suggests that KANs' strong function approximation capabilities are particularly advantageous when handling limited and noisy sensor data. In contrast, when multiple IMU sensors are available, the improvement margin diminishes. Specifically, in the OPPO dataset, which includes 77-channel inputs from IMUs placed on multiple body locations, the KAN-MLP-Mixer and MLPHAR models achieved comparable performance. Nevertheless, substantial gains were consistently observed when only a single accelerometer or IMU sensor served as input. A plausible explanation is that richer sensing mitigates the limitations of the MLPHAR model, while the complexity introduced by numerous input channels and diverse body placements reduces the effectiveness of the KAN-based embedding. Overall, these results validate the robustness of the proposed hybrid design across varying sensor configurations and underscore its practical value for real-world HAR deployments with heterogeneous sensing conditions.

\subsection{Extending Hybrid Design Across Diverse Window Size}

\begin{figure*}[hbt]
\centering
\includegraphics[width=\linewidth]{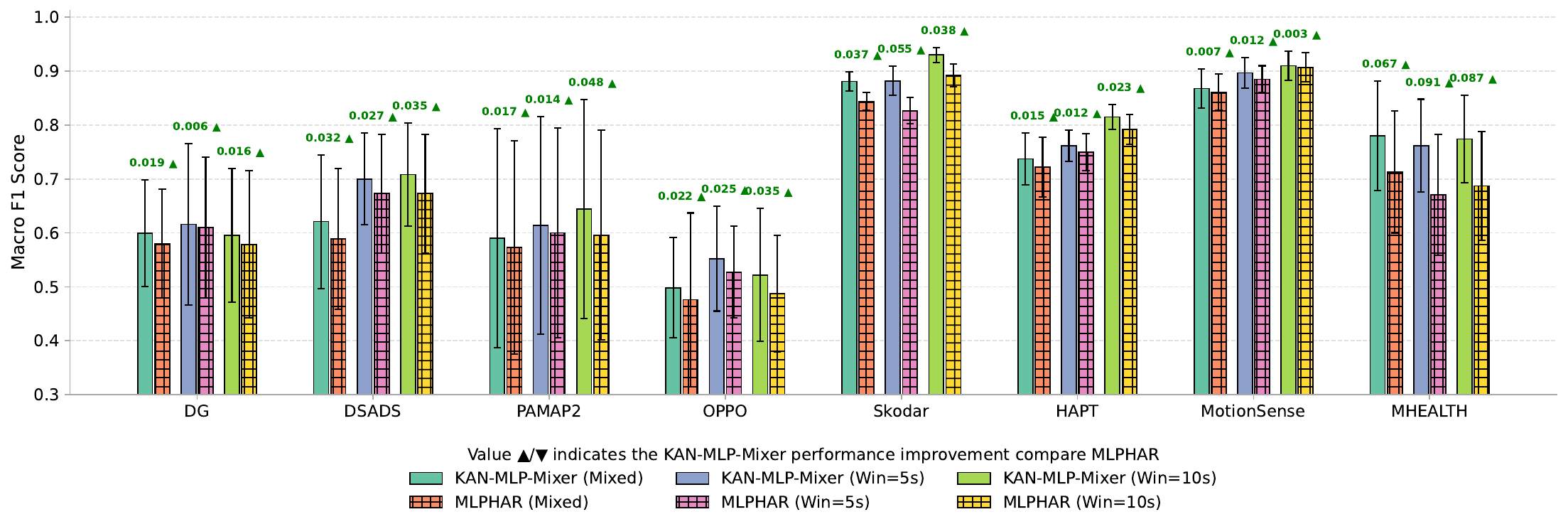}
\caption{Performance comparison for KAN-MLP-Mixer and MLPHAR models under three window size configurations. Numerical annotations show the absolute performance difference between models for each condition; $\blacktriangle$ favors KAN-MLP-Mixer, $\blacktriangledown$ favors MLPHAR.\textbf{ Across eight datasets, the KAN-MLP-Mixer model achieved an average macro F1 score improvement of 4.14\% with a mixed window size, 4.38\% with a 5-second window size, and 5.39\% with a 10-second window size, compared to the MLPHAR baseline.}}
\label{fig:result_cross_window}
\end{figure*}

To further assess the flexibility of the proposed KAN-MLP hybrid architecture, we investigated its performance under varying temporal window sizes. 
The data from a single IMU was used in this experiment.
We select the sensors as shown in \cref{tab:datasets}.
Specifically, we compared three configurations: (1) activity-oriented mixed window sizes from 1 to 5 seconds as shown in \cref{tab:datasets}. (2) uniform 5-second windows, and (3) uniform 10-second windows. The evaluation was conducted across eight datasets, as illustrated in \cref{fig:result_cross_window}.

Across all datasets, the KAN-MLP-Mixer consistently outperformed the MLPHAR baseline, regardless of the windowing strategy. On average, the KAN-MLP-Mixer achieved a macro F1 score improvement of 4.14\% with a mixed window size, 4.38\% with a 5-second window size, and 5.39\% with a 10-second window size. These results demonstrate that the hybrid model not only adapts well to activity-driven heterogeneous windowing across different datasets but also benefits from longer uniform windows, which offer more temporal context.
The largest gains were observed under the 10-second window condition, suggesting that the KAN-MLP-Mixer architecture is particularly effective at leveraging richer temporal information to improve classification accuracy. The robustness of the model across different window sizes further highlights its practical applicability to real-world scenarios, where sensing and segmentation strategies may vary depending on deployment constraints.
Overall, these results reinforce the versatility and strong generalization ability of the KAN-MLP hybrid design across diverse temporal settings.

\subsection{Extending Hybrid Design Across Diverse Neural Backbones}

To evaluate the generalizability of the proposed hybrid design, we further extended our modular approach, specifically, the use of \textbf{KAN-based embedding layers} and \textbf{LarctanKAN classifiers}, to a set of widely adopted non-pure MLP-based neural network architectures, including \textbf{MCNN}, \textbf{DeepConvLSTM}~\cite{ordonez2016deep}, and the lightweight \textbf{TinyHAR}~\cite{zhou2022tinyhar} models. The MCNN model comprises four convolutional layers for feature extraction, followed by two fully connected layers serving as the classifier. DeepConvLSTM shares a similar feature extraction and classification structure with MCNN but incorporates an additional LSTM module between the feature extractor and the classifier to capture temporal dependencies. TinyHAR features a more sophisticated and efficient design, integrating transformer-based and MLP-based cross-channel feature interaction and fusion modules, in addition to convolutional layers for feature extraction, LSTM layers for temporal modeling, and MLP layers for final classification. Together, these three models encompass a wide range of state-of-the-art components commonly used in deep learning-based HAR studies, making them ideal candidates for evaluating the generalizability performance of KANs-based hybrid models. In each model, we define the convolutional blocks as the feature extractor, the MLP layers at the end as the classifier, and the remaining modules between the feature extractor and classifier as the backbone. Notably, the MCNN model, with its simpler architecture, does not contain a distinct backbone module.

\begin{figure*}[hbt]
\centering
\includegraphics[width=0.99\linewidth]{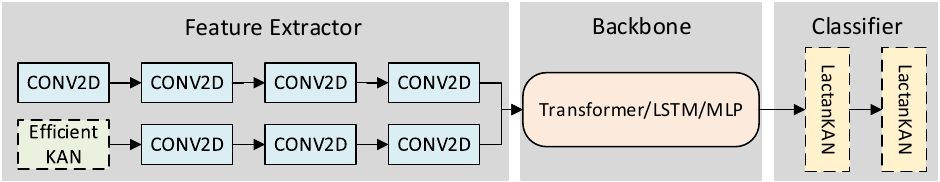}
\caption{The extending hybrid design across diverse neural backbones, the original models only have the pure convolutional layers on the feature extractor and the MLP layer on the classifier. We tested several combinations by replacing the first convolutional layer in the feature extractor or MLPs in the classifier with KANs. The combination shown above is denoted as \textbf{K-B-K}.}
\label{fig:hybrid_architecture}
\end{figure*}

In these experiments, we used the sensor configurations summarized in \cref{tab:datasets}. Specifically, we replaced the original first convolutional layer in the feature extraction block with \textbf{EfficientKAN} module, where the linear kernel function in the convolutional layer is replaced by the EfficientKAN, while preserving the backbone design of each model. Similarly, we substituted their output classifier heads with \textbf{LarctanKAN} modules. This modular insertion allowed us to assess whether our hybrid philosophy, originally optimized for pure-MLP models, could transfer effectively to more complex backbones commonly deployed in HAR tasks. \cref{fig:hybrid_architecture} illustrates the resulting hybrid architectures. In the original model designs, there was typically a single convolutional branch; in our hybrid configuration, we split this branch into two parallel branches, introducing a KAN layer into one of them. Each branch outputs half the number of channels compared to the original single-branch configuration, thereby maintaining the same overall output dimensionality before feeding into the backbone. We compare our hybrid architecture with a version where the KAN block is replaced by a vanilla CNN block, matching the other branch.

We evaluated three variants based on different combinations:
\begin{itemize}
    \item \textbf{K-B-M:} EfficientKAN in the first convolutional layer, original backbone, and MLP classifier.
    \item \textbf{C-B-K:} Original CNN-based feature extractor and backbone, LarctanKAN classifier.
    \item \textbf{K-B-K:} EfficientKAN in the first convolutional layer and LarctanKAN classifier, and original backbone, as shown in \cref{fig:hybrid_architecture}.
\end{itemize}

\begin{table}[ht]
\centering
\caption{Performance of hybrid variants across diverse neural backbones among 8 datasets (macro-F1 ± std).}
\label{tab:multimodel-results}
\scriptsize
\begin{tabular}{lccccccccc}
\toprule
\textbf{Model} & \textbf{DG} & \textbf{DSADS} & \textbf{PAMAP2} & \textbf{OPPO} & \textbf{Skodar} & \textbf{HAPT} & \textbf{MotionSense} & \textbf{MHEALTH} & \textbf{Gain (\%)} \\
\midrule

K-B-M & 0.654±0.164 & 0.657±0.089 & 0.615±0.227 & 0.544±0.140 & 0.964±0.008 & 0.819±0.024 & 0.914±0.032 & 0.642±0.151 & +0.24 \\
C-B-K & 0.668±0.148 & 0.675±0.090 & 0.615±0.204 & \textbf{0.575±0.135} & 0.960±0.014 & 0.821±0.027 & 0.919±0.034 & 0.652±0.143 & +1.77 \\
\textbf{K-B-K} & \textbf{0.672±0.159} & \textbf{0.686±0.080} & \textbf{0.616±0.196} & 0.564±0.122 & \textbf{0.968±0.009} & \textbf{0.825±0.025} & 0.919±0.035 & \textbf{0.659±0.137} & \textbf{+2.13} \\
MCNN & 0.659±0.155 & 0.657±0.091 & 0.607±0.202 & 0.567±0.146 & 0.952±0.012 & 0.817±0.030 & \textbf{0.921±0.030} & 0.615±0.142 & 0.00 \\
\hline
\hline
K-B-M & \textbf{0.680}±0.147 & 0.683±0.098 & 0.631±0.216 & 0.593±0.156 & 0.964±0.013 & 0.834±0.029 & 0.920±0.024 & 0.635±0.153 & +0.22 \\
C-B-K & 0.638±0.158 & 0.678±0.100 & 0.629±0.216 & \textbf{0.599}±0.152 & \textbf{0.971}±0.005 & \textbf{0.837}±0.025 & 0.929±0.029 & \textbf{0.684}±0.157 & +0.61 \\
\textbf{K-B-K} & 0.650±0.155 & \textbf{0.689}±0.082 & \textbf{0.654}±0.214 & 0.596±0.157 & 0.959±0.016 & 0.830±0.032 & \textbf{0.936}±0.021 & 0.671±0.179 & \textbf{+1.09} \\
DeepConvLSTM & 0.669±0.150 & 0.677±0.106 & 0.614±0.243 & 0.586±0.157 & 0.966±0.015 & 0.838±0.020 & 0.925±0.034 & 0.656±0.147 & 0.00 \\

\hline
\hline
K-B-M & \textbf{0.665}±0.151 & \textbf{0.670}±0.074 & 0.624±0.207 & 0.589±0.140 & \textbf{0.955}±0.013 & 0.810±0.029 & \textbf{0.928}±0.017 & 0.580±0.146 & -0.16 \\
C-B-K & 0.641±0.176 & 0.655±0.141 & \textbf{0.643}±0.213 & 0.561±0.140 & 0.940±0.011 & 0.810±0.041 & 0.914±0.021 & 0.615±0.134 & -0.82 \\
\textbf{K-B-K} & 0.631±0.174 & 0.636±0.085 & 0.634±0.200 & \textbf{0.590}±0.129 & 0.931±0.028 & \textbf{0.814}±0.029 & 0.910±0.029 & \textbf{0.689}±0.109 & \textbf{+0.46} \\
TinyHAR & 0.658±0.168 & 0.659±0.129 & 0.623±0.211 & 0.589±0.129 & 0.930±0.014 & 0.806±0.035 & 0.913±0.023 & 0.636±0.119 & 0.00 \\
\bottomrule
\end{tabular}
\end{table}

\cref{tab:multimodel-results} presents the performance summary of the hybrid variants neural network across the three models, it can be found that the K-B-K models achieve a better macro F1 score than the original models and other variants. 
Besides, the average performance revealed a clear trend: KAN-based modules delivered the largest performance gains (2.13\%) in the simplest architecture (MCNN), with progressively smaller improvements (1.09\%) for DeepConvLSTM and especially for the advanced TinyHAR (0.46\%). 
This indicates that the benefits of the EfficientKAN embedding and LarctanKAN classifier are inversely related to the capacity and sophistication of the host architecture. In other words, the more complex and expressive the backbone, the less additional value the KAN components seem to provide.

One fundamental reason for the diminishing returns is the difference in model capacity across these backbones. MCNN, being a relatively shallow CNN-only model, has limited ability to capture complex, non-linear relationships in sensor data.
It relies on local convolutional filters and static activation functions, so augmenting it with KAN modules significantly boosts its expressive power.
In contrast, DeepConvLSTM already possesses greater representational capacity by combining convolutional feature extractors with LSTM layers that capture temporal dynamics.
This means the baseline DeepConvLSTM can already model many of the patterns that KAN would help with in a simpler network. Any additional complex mapping provided by the KAN embedding or classifier might partially duplicate what the CNN+LSTM is learning (a form of learning redundancy), resulting in only moderate gains. Finally, TinyHAR is an even more sophisticated model, incorporating not just convolution and recurrent layers but also attention-based mechanisms (transformer encoder blocks) for cross-channel feature interaction.
Such attention modules effectively perform a data-dependent re-weighting of features, which is conceptually similar to the kernel-based weighting that KAN modules implement. Thus, the TinyHAR baseline is already extremely expressive – it was designed to match or exceed DeepConvLSTM’s accuracy with a much smaller parameter count by cleverly maximizing model utilization.
There is simply less “unused capacity” for KAN to fill in TinyHAR. The small improvement observed in TinyHAR with KAN likely comes from fine-tuning the decision boundaries or subtle feature tweaks, but much of KAN’s powerful function-approximation ability is redundant given TinyHAR’s existing components. In essence, as the backbone’s capacity to approximate complex functions increases, the marginal utility of inserting KAN layers decreases.

\subsection{Evaluating Parameter and Computational Efficiency of Hybrid KAN Designs}
\label{sec: evaluate_efficiency}

\begin{figure*}[hbt]
\centering
\includegraphics[width=\linewidth]{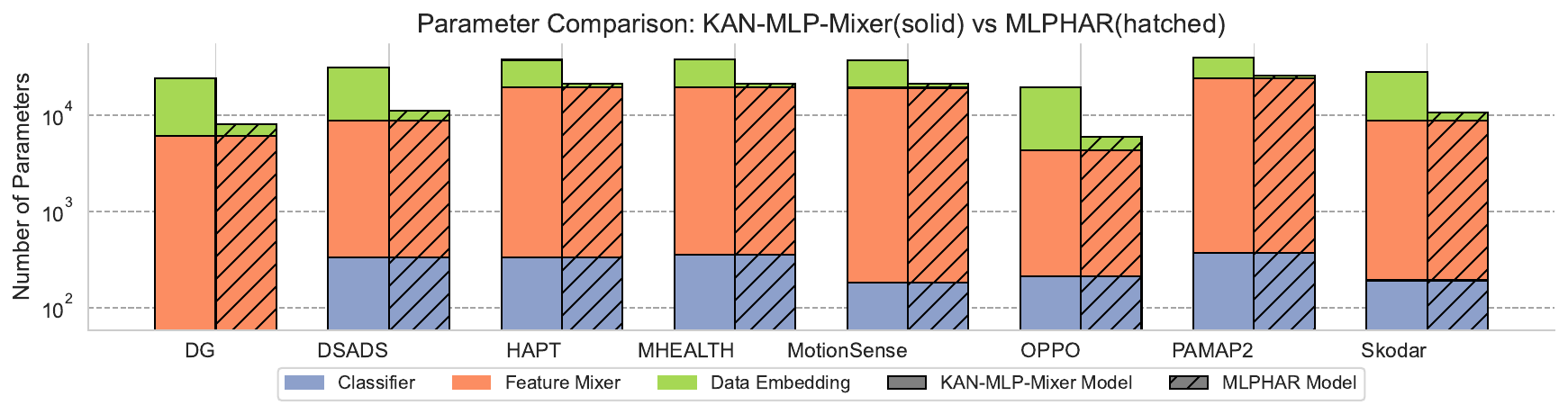}
\caption{Number of parameters in different components of KAN-MLP-Mixer and MLPHAR models.(The \textbf{LarctanKAN} based classifier and MLP modules have the same number of trainable parameters; therefore, the parameter difference between the \textbf{KAN-MLP-Mixer} and the \textbf{MLPHAR} models originates entirely from the data embedding module.)}
\label{fig:paramter_comp}
\end{figure*}

We further assessed the practicality and efficiency of the hybrid KAN-MLP-Mixer architecture by analyzing its parameter and computational complexities relative to the baseline MLPHAR model. As depicted in \cref{fig:paramter_comp}, when the KAN-MLP-Mixer and MLPHAR models have identical layer configurations and input/output dimensions, the hybrid KAN design generally incurs a higher total parameter count. This increase arises primarily because each KAN layer inherently introduces approximately \textit{G} times more parameters than a corresponding MLP layer, where \textit{G} is the number of grid points used in the B-spline basis. However, by strategically assigning KAN modules only to the data embedding and classifier components, and preserving MLP in the typically larger feature mixer module, the hybrid design achieves a balanced trade-off between parameter efficiency and expressive capacity.

\begin{figure*}[hbt]
\centering
\includegraphics[width=\linewidth]{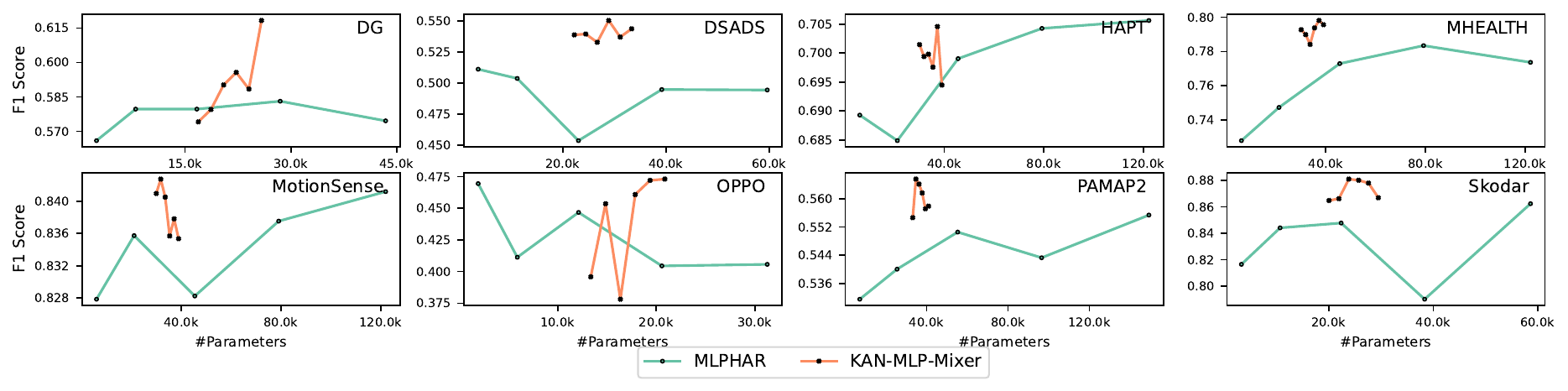}
\caption{Parameter-efficiency when scaling models comparing between KAN-MLP-Mixer and MLPHAR across eight benchmark datasets. For KAN-MLP-Mixer, the grid size $G$ was systematically increased from 1 to 6; for MLPHAR, the hidden dimensions in the data embedding module were varied from 8, 16, 24, 32, to 40. These results illustrate the scalability of KAN-hybrid models and their ability to achieve competitive or superior performance compared to MLP-only models under varying parameter budgets.}
\label{fig:f1_vs_para}
\end{figure*}

In order to investigate the effect of model size on performance, we also increased the size of the data embedding in MLPHAR models 
(see \cref{fig:f1_vs_para}) to elucidate whether our gain in performance can be matched by simply increasing the size of MLP-only models. This parameter-efficiency comparison shows that, across multiple benchmark datasets, the KAN-MLP-Mixer consistently outperforms or matches the performance of the MLPHAR baseline while using fewer or comparable parameters. Specifically, as the grid size \textit{G} increased from 1 to 6 in the KAN-based components and the hidden dimensions varied from 8 to 40 in the data embedding module from MLPHAR, the hybrid model demonstrated superior scalability. In datasets such as DG, DSADS, and HAPT, the KAN-MLP-Mixer achieved notably higher macro F1 scores even at lower parameter counts compared to the MLPHAR baseline. This suggests that the hybrid model’s strategically placed KAN layers leverage their powerful function approximation capabilities to learn richer representations, leading to improved accuracy without excessively increasing model size.

\begin{figure*}[hbt]
\centering
\includegraphics[width=\linewidth]{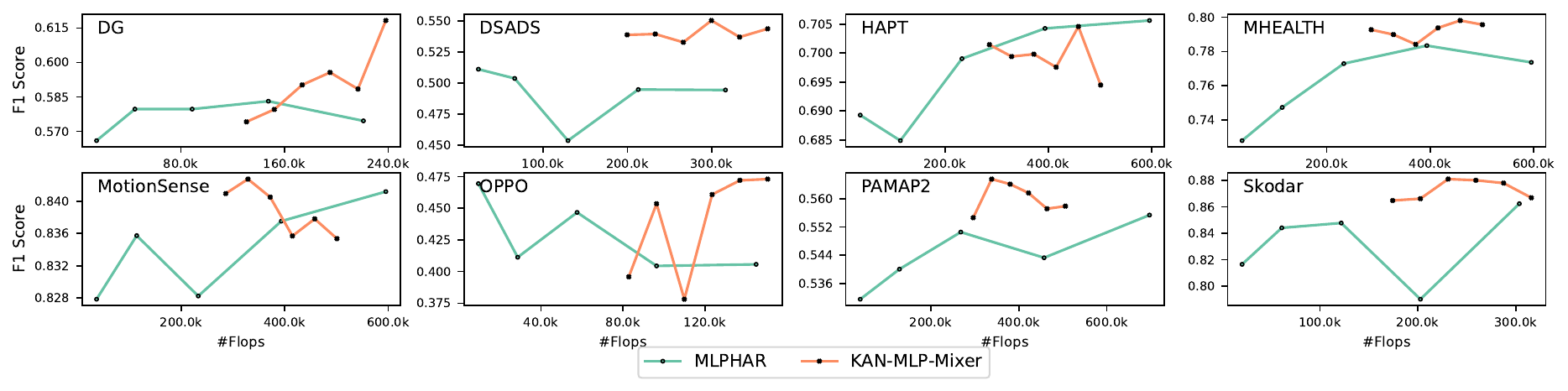}
\caption{Computational efficiency comparison between KAN-MLP-Mixer and MLPHAR across eight benchmark datasets. For KAN-MLP-Mixer, grid size $G$ was increased from 1 to 6; for MLPHAR, hidden dimensions were varied from 8, 16, 24, 32, to 40. These results demonstrate how KAN-based models maintain competitive or superior performance relative to MLP-based models while offering favorable computational profiles under varying FLOP budgets.}
\label{fig:f1_vs_flops}
\end{figure*}

Computational efficiency analysis further corroborates the practicality of the hybrid approach, as shown in \cref{fig:f1_vs_flops}. Although KAN layers typically demand higher computation due to the evaluation of spline-based activation functions, the KAN-MLP-Mixer still maintains competitive or superior computational efficiency across the benchmark datasets. Notably, in datasets like DG, DSADS, OPPO, and Skodar, the hybrid architecture achieves higher macro F1 scores at comparable or lower floating-point operation (FLOP) budgets relative to MLPHAR. This efficiency arises primarily because the KAN layers’ ability to accurately model complex input-output relationships with fewer neurons and layers compensates for their inherently higher per-layer computational demands. Consequently, the hybrid architecture can provide enhanced performance within practical computational constraints, making it particularly suitable for deployment in resource-constrained wearable and ubiquitous computing scenarios.

\subsection{Limitations of the Current Study}

Despite the promising results and comprehensive evaluation presented, this study is subject to several limitations that should be acknowledged:

\textbf{Dataset Scope and Generalization.}
Our evaluation relies primarily on publicly available benchmark datasets, which—while diverse in sensor modalities, positions, and activity complexity—may not encompass the full variety and unpredictability of real-world HAR scenarios. Many real-world datasets include intermittent sensor noise, missing data, and user-specific variations not fully captured by the controlled benchmark conditions.

\textbf{Sensor Modality and Placement Constraints.}
The selected datasets primarily utilize accelerometers, gyroscopes, and occasionally magnetometers. Other sensor modalities, such as physiological sensors (e.g., heart rate monitors, electromyography), environmental sensors (e.g., ambient light or temperature), or fusion with video and audio data, were not considered. Consequently, the performance of the proposed architecture in multi-modal sensing contexts remains unexplored.

\textbf{Computational and Energy Constraints.}
While the proposed KAN-MLP-Mixer architecture significantly improves model performance, its practical deployment on resource-constrained devices (e.g., low-powered wearables or IoT sensors) has not been directly validated. The trade-off between improved accuracy and increased computational complexity or energy consumption—common challenges in real-world applications—requires further investigation.

\textbf{Architectural and Hyperparameter Optimization.}
Our proposed model utilizes specific variants of KANs (EfficientKAN and LarctanKAN) selected based on empirical performance. The sensitivity of our results to hyperparameters such as KAN spline complexity, grid size, or the number of layers and neurons remains to be fully quantified. Additionally, automated methods for systematically selecting or optimizing these parameters have yet to be explored.

\subsection{Future Work Directions}

Based on the insights gained from this study, several promising research directions emerge:

\textbf{Extension to Multi-modal HAR Systems}.
Exploring the integration of additional sensor modalities beyond inertial sensors—such as physiological, audio, visual, and environmental data—offers an intriguing avenue for enhancing HAR performance and robustness. Investigating how KAN-based architectures perform within such multi-modal data fusion frameworks could further validate their utility in practical HAR deployments.

\textbf{Resource-aware Optimization for Edge Devices}.
Evaluating and optimizing the proposed hybrid architecture explicitly for computational and energy efficiency on edge and wearable devices represents an important next step. Future work should examine model compression techniques, efficient quantization, lightweight model structures, and energy-aware training procedures to ensure the hybrid model remains practical for ubiquitous deployment.

\textbf{Automated Architecture Search and Hyperparameter Optimization}.
Investigating automated techniques, such as Neural Architecture Search (NAS), to systematically identify optimal KAN placements, architectural parameters, and hyperparameter settings would further refine performance. Additionally, rigorous sensitivity analyses of hyperparameters could provide clearer guidelines for deploying KAN-based hybrid architectures in diverse scenarios.

\textbf{Interpretable and Explainable KAN-based Models}.
Given the inherent interpretability of KAN modules, future research should investigate interpretability-focused evaluations, providing clearer insights into what sensor-derived features and patterns contribute most significantly to accurate HAR predictions. Enhanced interpretability could substantially increase user trust and acceptance in critical application domains such as healthcare or rehabilitation.

% In conclusion, addressing these limitations and exploring these directions will significantly enhance the practical viability, robustness, and generalizability of hybrid KAN-MLP architectures in ubiquitous computing environments.

\subsection{Design Guidelines for Sensor-Based HAR}

Our comprehensive ablation results, supported by theoretical insights and synthetic function fitting experiments, offer actionable design principles for integrating KANs into modern neural architectures. We distill these findings into three practical guidelines:

\begin{enumerate}
    \item \textbf{Use KANs selectively, not universally:}
    While KANs demonstrate strong representational capacity—particularly for modeling smooth, continuous functions—our findings show that \textbf{full-scale replacement of MLPs with KANs leads to significant performance degradation}, especially in deeper layers. Instead of wholesale substitution, KANs should be applied strategically where their strengths (e.g., spline-based approximation, localized activation) align with the structure of the input data.

    \item \textbf{Match model component to signal characteristics:}
    Our empirical results suggest a strong link between model layer type and the nature of the signal it processes:
    \begin{itemize}
        \item Use \textbf{KANs for data embedding} when inputs are smooth and periodic, such as IMU sensor data in HAR.
        \item Use \textbf{MLPs for latent feature mixing}, where signal continuity is less pronounced and stability, generalization, and scalability are crucial.
        \item Use \textbf{LarctanKANs for classification}, where stable, bounded nonlinearities can model decision boundaries without inducing overshooting or noise amplification.
    \end{itemize}

    \item \textbf{Prioritize hybrid modularity over architectural novelty:}
    The key to successful KAN integration lies not in the novelty of the model itself, but in the \textbf{modular hybridization of architectures}. Each module should be chosen and positioned based on its demonstrated strengths. Our EfficientKAN–MLP–LarctanKAN configuration provides a template for how to combine expressive, stable, and efficient modules into a unified architecture—one that outperforms both pure MLP and pure KAN-based designs across diverse datasets.
\end{enumerate}

These guidelines are intended to help researchers and practitioners move beyond one-size-fits-all architectures, encouraging principled design choices based on signal properties, module behavior, and empirical evidence. We believe this modular mindset will be essential as the field continues to integrate emerging neural operators into scalable, interpretable, and domain-adaptive machine learning systems.

\section{Conclusion}

This work systematically explored integrating KANs into robust and computationally efficient MLP based architectures for IMU-based HAR. Recognizing KANs' high expressivity yet sensitivity to noisy, real-world data and MLPs' robustness and efficiency, we introduced the KAN-MLP-Mixer hybrid architecture. This hybrid design strategically employs EfficientKAN for adaptive input embedding, retains standard MLP layers for intermediate feature mixing, and incorporates a specialized LarctanKAN classifier. Evaluations across eight diverse HAR datasets confirmed the effectiveness of this targeted integration, achieving an average macro F1 score improvement of 5.33\% over the pure-MLP baseline (MLPHAR). Our results clearly demonstrate that selective hybridization significantly surpasses both standalone KAN and MLP models. Furthermore, extending this hybrid strategy to other state-of-the-art neural backbones consistently improved their performance, underscoring its broad applicability. This work thus provides concrete guidelines for effectively leveraging KANs in practical wearable sensing scenarios, marking a promising advancement towards accurate HAR systems harnessing the potential function approximation power of KANs.

% \begin{acks}
% \end{acks}

\bibliographystyle{ACM-Reference-Format}
\bibliography{sample-base}

\end{document}